\documentclass[twocolumn]{article}

\usepackage{PRIMEarxiv}
\usepackage[utf8]{inputenc}
\usepackage[T1]{fontenc}
\usepackage{hyperref}
\usepackage{url}
\usepackage{booktabs}
\usepackage{amsfonts}
\usepackage{nicefrac}
\usepackage{microtype}
\usepackage{fancyhdr}
\usepackage{threeparttable}
\usepackage{graphicx}
\usepackage{lipsum}
\usepackage[authoryear,round]{natbib}  
\usepackage{hyperref}
\usepackage{caption}

\graphicspath{{media/}}

\PassOptionsToPackage{hyphens}{url}
\usepackage{hyperref} 

\hypersetup{
  breaklinks=true,
}

\title{California Wildfire Inventory (CAWFI): \newline
An Extensive Dataset for Predictive Techniques based on Artificial Intelligence}

\author{
  Rohan Tan Bhowmik$^{1,*}$, Youn Soo Jung$^{1,2}$, Juan Aguilera$^{1,3}$, Mary Prunicki$^{1,2}$, Kari Nadeau$^{1,2}$ \\
  \\
  $^1$ Stanford University, Stanford, California \\
  $^2$ Harvard University T. H. Chan School of Public Health, Cambridge, Massachusetts \\
  $^3$ University of Texas Health Science Center, Houston, Texas \\
  $^*$\texttt{rbhowmik@stanford.edu} \\
}

\begin{document}

\twocolumn[
\begin{@twocolumnfalse}

\maketitle

\vspace{-1cm}

\begin{abstract}
Due to climate change and the disruption of ecosystems worldwide, wildfires are increasingly impacting environment, infrastructure, and human lives globally. Additionally, an exacerbating climate crisis means that these losses would continue to grow if preventative measures are not implemented. Though recent advancements in artificial intelligence enable wildfire management techniques, most deployed solutions focus on detecting wildfires after ignition. The development of predictive techniques with high accuracy requires extensive datasets to train machine learning models. This paper presents the California Wildfire Inventory (CAWFI), a wildfire database of over 37 million data points for building and training wildfire prediction solutions, thereby potentially preventing megafires and flash fires by addressing them before they spark. The dataset compiles daily historical California wildfire data from 2012 to 2018 and indicator data from 2012 to 2022. The indicator data consists of leading indicators (meteorological data correlating to wildfire-prone conditions), trailing indicators (environmental data correlating to prior and early wildfire activity), and geological indicators (vegetation and elevation data dictating wildfire risk and spread patterns). CAWFI has already demonstrated success when used to train a spatio-temporal artificial intelligence model, predicting 85.7\% of future wildfires larger than 300,000 acres when trained on 2012--2017 indicator data. This dataset is intended to enable wildfire prediction research and solutions as well as set a precedent for future wildfire databases in other regions.
\end{abstract}


\vspace{0.75cm}

\end{@twocolumnfalse}
]

\section{Introduction}
\label{sec:introduction}
Wildfires are an essential component of and serve critical roles in ecological and biodiversity balance. These natural processes, which dictate the cycling of vegetation, nutrients, water, and heat, are a crucial component of maintaining healthy ecosystems \cite{Keeley2009, DeBaNO2000}. However, climate change brought about contemporary issues including increasing global temperatures, prolonged droughts, and shifting weather patterns, which have intensified the frequency, duration, and severity of wildfires. 

Additionally, other manmade causes of wildfire, including rapid growth of wildland-urban interface, rapid housing development, and infrastructure failures, have further exacerbated wildfire risks \citep{Radeloff2018, Macomber2024}. Excessive and intense wildfires disrupt this cycle, leading to long-term ecological damage as well as impacts to life and quality-of-life.

Despite significant effort dedicated to prescribed burns, fire suppression, and other means of wildfire management, U.S. wildfires have burnt more than 206 million acres and cost the United States Forest Service and Department of the Interior \$48.7 billion in wildfire management since 1983 \citep{Bishop2023}. Although the annual wildfire incident count in the U.S. has remained relatively constant over the last 4 decades, the acreage burnt has more than tripled while annual suppression costs have increased by roughly 15 times in 2022 as compared to 1985 \citep{NIFC2023}. In 2023, devastating wildfire outbreaks ravaged areas such as Hawaii, Canada, Chile, and Europe, leading to hundreds of lost lives and billions in rebuilding costs \citep{Sinco2023,Budner2023,CIFFC2023,Slaybaugh2024,Yoder2024}. More than ever, the world needs easily deployable solutions to predict major wildfires, identify wildfires at their inception, and minimize the impact of wildfires that otherwise would significantly endanger the environment and communities.

Historically, California’s dry summers and hotter weather made its forests more susceptible to wildfires; more significantly, though, a legacy of heightened wildfire suppression and constantly disrupting their balance contributed to an unusually large buildup of dry unburned fuel \citep{Syphard2007}. Combined with a recent history of severe droughts, ignitions of these biomass in the past few years have created some of the most devastating and deadly blazes in modern history. Those such as the Dixie and August Complex Fires burned through many acres, razing nutrients from the soil while dispersing noxious gases and particulates into lungs throughout the nation \citep{CarrerasSospedra2024}. Others such as the Camp Fire led to tremendous loss of structures and lives through their speed \citep{Maranghides2023}. These circumstances make California a prime region to study and develop strategies for managing virulent fires.

This paper details the creation of the California Wildfire Inventory (CAWFI) wildfire prediction database, and discusses its potential applications. The wildfire database developed in this work contains over 37 million data points, aggregating catalogs of historical wildfires in California between 2012-2018 and meteorological, and geological data between 2012-2022. Building on the authors’ prior work on predicting wildfires using machine learning networks trained on these data, this database is developed with the intention to serve as a resource for future wildfire prediction research utilizing artificial intelligence (AI) techniques based on machine learning (ML) algorithms \citep{Bhowmik2023}.

\section{Background and Motivation}
\label{sec:background}
Over the past few years, artificial intelligence techniques have improved exponentially in efficiency and robustness, transforming numerous domains of science and technology. Specifically, the advent of convolutional neural networks (CNN), recurrent neural networks (RNN), transformer networks, and related developments in AI architectures in conjunction with burgeoning datasets that have enabled robust training of large AI models have supercharged the fields of computer vision, image processing, and natural language processing \citep{LeCun2015,Vaswani2017,Russell2022}. Recent applications such as ChatGPT, Midjourney, and MusicLM have thrust AI’s potential into the public eye, further accelerating innovations and applications in various domains \citep{OpenAI2022,Midjourney2023,Google2023}. The potential applications of AI in predicting and managing wildfires have also been explored \citep{Sathishkumar2023,Bouguettya2022,Bhowmik2023,Huot2022}.

There are two major thrusts in approaching wildfire management. Wildfire detection techniques focus on early detection after they have already started. This approach aims to identify newly-ignited wildfires and extinguish them before they spread and evolve into larger ones. On the other hand, wildfire prediction methods focus on allocating resources to high wildfires risk areas ahead of the occurrence of potentially large wildfire events. This approach aims to identify high risk areas and deploy resources to prevent the wildfire from burning in the first place. A summary and comparison of notable wildfire detection and prediction solutions are summarized in Table ~\ref{table:1}. 

A number of different AI solutions for addressing wildfires are beginning to be deployed and tested in the past few years. For example, FireScout rents its services to over a thousand government cameras across the Western United States \citep{FireScout2022}. Using AI-based image processing techniques, the cameras identify wildfires by detecting unusual smoke plumage. Once the wildfires are identified, the FireScout system alerts the fire fighting authorities immediately. As another example, Dryad sells its solar-powered Silvanet wildfire sensors, which monitor hydrogen, carbon monoxide, and other gasses emitted by forest fires \citep{Dryad2021}. Using AI, a network of sensors 100 meters apart will detect wildfires within the first hour of pyrolysis. Additionally, many peer-reviewed papers have contributed toward advancing wildfire detection accuracy. Some recent techniques include: pre-trained models to detect fire and smoke from visible-light images \citep{Sathishkumar2023}; regression models to classify irregularities in moderate resolution imaging spectroradiometer (MODIS) readings as wildfires \citep{Ding2023,GISGeography2023}; and unmanned aerial vehicles employing computer vision algorithms on optical and thermal infrared camera measurements \citep{Bouguettya2022}. 

However, relying solely on wildfire detection techniques does not allow a wildfire to be managed at its onset. These detection techniques rely on identifying wildfires that have been burning for a period of time. The additional time required for firefighting crews to arrive on scene could translate to hours or days during which a wildfire could devastate forests and spread uncontrollably. In the worst-case scenarios such as the notoriously devastating 2018 Camp Fire in Northern California, flames engulfed over 150,000 acres in just the first four hours \citep{CalFire2022}. High-speed winds and burning infrastructure hampered first responders, air resources, and evacuation efforts from quelling the fire and mitigating loss of life any sooner. When wildfires such as Camp Fire can spread at such a rapid pace, the aforementioned detection techniques fall short of wholly containing megafires. 

\begin{table*}[t]
\captionsetup{font=small}
\caption{Summary of notable wildfire detection and prediction techniques, comparing methodology, achievements, and limitations. Abbreviations: CCTV = closed-circuit television; MODIS = Moderate Resolution Imaging Spectroradiometer; AI = artificial intelligence. The F1 score is the harmonic mean of specificity and sensitivity.}
\label{tab:summary-wildfire-techniques}
\centering
\footnotesize
\begin{threeparttable}
\begin{tabular}{@{}p{1.0cm}p{2.0cm}p{4.6cm}p{3.76cm}p{3.4cm}@{}}
\toprule
\textbf{Approach} & \textbf{Reference / Product} & \textbf{Method and Equipment Used} & \textbf{Merit and Achievements} & \textbf{Limitations} \\
\midrule
\addlinespace[2pt]
Detection & FireScout (2022) & Live CCTV network with AI image processing (smoke plume detection) & $>$99\% accuracy within 1 minute of first smoke; covers broad forested areas & High installation/maintenance cost; rapid wildfires still hard to mitigate using detection alone \\
\addlinespace[2pt]
Detection & Silvanet (Dryad, 2021) & Solar-powered air-quality sensor mesh (H$_2$, CO, etc.) & Guaranteed detection within 1 hour; self-sufficient (10–15 yr lifetime) & Dense mesh required; costly for large-scale deployments \\
\addlinespace[2pt]
Detection & Sathishkumar et al. (2023) & Transfer learning on visible-light imagery (fire/smoke classification) & 98.72\% accuracy; F1 = 0.973; detects smoke and/or fire & Limited to ongoing fires (post-ignition) \\
\addlinespace[2pt]
Detection & Ding et al. (2023) & Regression on satellite MODIS brightness temperature & RMSE = 0.0730; widely deployable with accessible live data & MODIS revisit $\sim$2 days; limited for real-time response \\
\addlinespace[2pt]
Detection & Bouguettaya et al. (2022) & UAVs with optical \& thermal IR; CV algorithms & Can leverage $>$99\% accuracy techniques; flexible coverage & Higher cost to scale and maintain over large areas \\
\addlinespace[2pt]
Prediction & Bhowmik et al. (2023) & Spatio-temporal AI on meteorological, environmental \& geological factors (this paper’s dataset) & Predicted 85.7\% of $>$300{,}000 acre fires up to two weeks in advance & California-only scope; comprehensive real-time feeds can be limited \\
\addlinespace[2pt]
Prediction & Huot et al. (2022) & ML using meteorological \& geological factors (MODIS) & Up to 35.7\% precision, 46.9\% recall (pixel-wise analysis) & MODIS revisit $\sim$2 days; focus on fire spread prediction \\
\addlinespace[2pt]
Prediction & Sykas et al. (2023) & EO4WildFires dataset (2018–2022) with meteorological drivers & Large multi-sensor, ML-ready benchmark; $\sim$32k wildfires across 45 countries & Benchmark dataset—downstream modeling performance varies by method \\
\addlinespace[2pt]
Prediction & Firemark AI (2023) & Geospatial AI with optical sensors \& UAVs & Claims 90\% accuracy up to 30 days in advance & Limited public details on data and methods \\
\bottomrule
\end{tabular}
\end{threeparttable}
\label{table:1}
\end{table*}

The second category of approaches is focused on wildfire predictions. These preventative solutions focus on allocating firefighting resources to areas of high wildfire risk. This method greatly mitigates the likelihood of megafires by addressing emerging wildfires in high risk areas at inception. Specifically, predicting when, where, and how hard wildfires may strike would allow for firefighting crews to remediate wildfire-prone conditions and/or be on standby to address new flames immediately. Recent solutions include a novel spatio-temporal AI architecture analyzing meteorological, environmental, and geological factors (utilizing this database) to predict wildfires up to two weeks in advance, and regression and random forest models on meteorological and MODIS data for projecting wildfire spread \citep{Bhowmik2023,Huot2022}. Datasets dedicated specifically for training wildfire prediction solutions include the multivariate “Next Day Wildfire Spread” dataset from the paper by Huot et al. and the large-scale ”EO4WildFires” dataset by Sykas et al. for predicting the severity of hypothetical wildfires in any selected area \citep{Huot2022,Sykas2023}. Skymount’s Firemark AI proposes the first commercial wildfire prediction solution with 90\% prediction accuracy up to 30 days in advance, though little about its data or methodology has been disclosed to the public \citep{Skymount2023}. 

Currently, wildfire data prediction techniques rely on a combination of remote sensors, surveillance footage, and satellite coverage for collecting data. Surveillance footage requires high installation and maintenance costs over large areas, with each acre requiring roughly 500 GB of footage memory and 27 kWh of camera energy annually \citep{KFConcept2023,Veesion2023}. Though manageable quantities over smaller areas such as corridors and rooms, scalability proves challenging over larger areas. Satellite data with moderate resolution imaging spectroradiometer instruments is commonly used in wildfire monitoring. This technique is good for long-term prediction applications, but it has limited use in real-time prediction due to the revisit time of satellites, which describes the time gap between satellite coverage of the same point on Earth as around 2 days. Of these data collection methods, remote meteorological and air quality sensor networks stand out as the most flexible, with each sensor covering multiple acres of land. As an example, the widely deployed PurpleAir sensor network measures a wide variety of air quality factors, transmits just 6.3 GB of data and consumes 8.8 kWh of power each year, which makes it more feasible to be implemented country-wide for wildfire prediction than surveillance or satellites \citep{PurpleAir2022}.

The database compiled and presented in this paper, dubbed California Wildfire Inventory, leverages existing remote sensor networks throughout California for collecting historical data as well as monitoring real-time data. Various indicators from sensor networks for large-scale wildfire prediction solutions are analyzed and compiled to address the current lack of widely-scalable wildfire prediction databases. In the long term, the goal is that CAWFI will aid and eventually expand with the global concerted efforts in systematically expanding sensor network coverage.

\section{Database Structure}
\label{sec:database}
This section outlines the methodology of selecting, sourcing, and processing different parameters accessible through remote sensor networks to construct CAWFI as a comprehensive wildfire database.
To predict wildfires in advance, indicators that directly impact wildfire conception and propagation must be collected and used to train machine learning models. These factors must be considered in conjunction with one another, as wildfires are driven by a complex interplay of ecological, meteorological, and geographical elements. By analyzing how these factors interact, predictive models can better estimate the likelihood and behavior of future wildfires. Thus, three types of wildfire indicators are considered: leading, trailing, and locational. Specifically, leading indicators demonstrate unusual trends or patterns in an area before a wildfire erupts, trailing indicators demonstrate trends during and right after wildfire ignition, and locational indicators affect the overall risk and spread of wildfires based on surrounding geographical characteristics.
The leading indicators include meteorological data. Trends in meteorological data indicate patterns in the weather and land conditions which reflect the risk of a wildfire igniting and raging into a conflagration. Meteorological data are leading indicators that can foretell the risk of lightning strikes, which cause more than 40\% of wildfires and 70\% acreage burnt in California \citep{Cart2023}. Regardless of the cause of ignition, these leading indicators can also determine the likelihood of and the speed at which a small kindle will exacerbate into a large wildfire. The meteorological leading indicators collected include:

\begin{enumerate}
  \item Temperature (measured in $^{\circ}\mathrm{C}$), which impacts levels of precipitation, water evaporation, etc.\ that contribute to fuel levels and wildfire-prone conditions.
  \item Dew point (measured in $^{\circ}\mathrm{C}$), an indirect measure of relative humidity that indicates air moisture; a lack thereof causes drier vegetation and escalated risk of wildfire occurrence.
  \item Wind speed (measured in knots/hour), which increases oxygen supply to combustible materials and accelerates fire spread.
\end{enumerate}

The trailing indicators are the environmental data. Trends in environmental data indicate gaseous emissions of smoke and nauseous gasses typically produced by the combustion of organic material. Specifically, irregularities in environmental data from areas within a forest indicate the beginnings of newly ignited wildfires. Over longer periods of time and larger areas, these trailing indicators can offer insight into the current progress and future trajectory of wildfire spread. As other non-wildfire activity can produce these irregular trailing indicators trends, environmental data should be compared with patterns in historical data along with meteorological leading indicators. The environmental trailing indicators collected include:

\begin{enumerate}
  \item $\mathrm{PM}_{2.5}$ (measured in $\mu$g/m$^{3}$), the concentration of particulate matter in the air with a diameter of 2.5\,$\mu$m or less \citep{Liu2016}.
  \item $\mathrm{PM}_{10}$ (measured in $\mu$g/m$^{3}$), the concentration of particulate matter in the air with a diameter of 10\,$\mu$m or less \citep{Liu2016}.
  \item $\mathrm{CO}$ (carbon monoxide) levels (measured in parts per million, ppm) \citep{Bela2022}.
  \item $\mathrm{NO}_{2}$ (nitrogen dioxide) levels (measured in parts per billion, ppb) \citep{Griffin2021}.
\end{enumerate}

The locational indicators include geological data. Geological data indicate land and forest conditions that dictate the speed and direction of a wildfire’s spread. These data include the flammability of an area’s vegetation as a factor of plants’ health and the terrain in which it lies. Although geological data remains relatively more constant over time as compared with the meteorological and environmental data, natural and human activity can significantly alter these geological factors over the span of years. The geological indicators collected include:

\begin{enumerate}
  \item Elevation (measured in m), with higher altitudes having drier air but lower temperatures; steeper uphill slopes accelerate fire spread.
  \item NDVI (Normalized Difference Vegetation Index, unitless), which quantifies vegetation productivity and health—less fertile plants are drier and more prone to ignition.
  \item Ecological factors (e.g., vegetation cover and fuel models) from LANDFIRE provide direct insights into biomass density and flammability \citep{LANDFIRE2024}.
\end{enumerate}

\begin{figure}[t]
  \centering
  \includegraphics[width=1.0\linewidth]{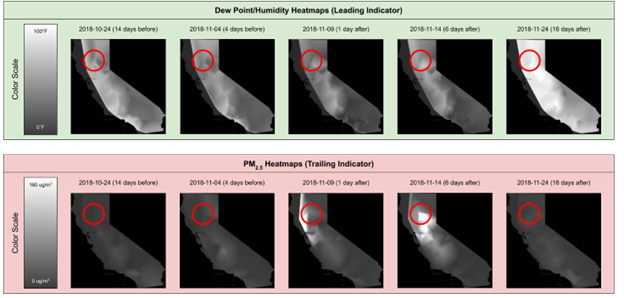}
  \caption{Selected dew point and $\mathrm{PM}_{2.5}$ heatmaps as examples of leading and trailing indicators. The red circle highlighted on each image highlights the location in which the Camp Fire erupted in California. The dew point heatmaps show the surrounding area having abnormally low humidity before the wildfire, while the $\mathrm{PM}_{2.5}$ graphs show the area having abnormally high readings during and after the wildfire. These trends provide the basis of predicting the risk of a wildfire igniting using meteorological and environmental factors.}
  \label{fig:1}
\end{figure}

As an example, in Figure ~\ref{fig:1}, dew point and $\mathrm{PM}_{2.5}$ measurements across California before, during, and after the 2018 Camp Fire are transformed and displayed as 8-bit images in which the pixels of brighter values correspond to areas with higher readings. On each heatmap, the area circled in red shows the epicenter of the Camp Fire near California’s Butte County. The dew point leading indicator heatmaps show lower humidity trends before the wildfire ignition, contributing to a drier climate more prone to wildfire spread. The $\mathrm{PM}_{2.5}$ displays abnormally high readings soon after Camp Fire’s ignition, only subsiding to lower values after the fire was contained. Including these example heatmaps, the leading meteorological indicators appear to demonstrate abnormal trends before the fire outbreak, and the trailing environmental indicators show abnormal trends after the outbreak (later than expected as the sensor is located farther from fire). As such, an AI network analyzing these heatmaps should be able to predict the Camp Fire before its outbreak. Indeed, our prior work has shown that when utilizing both leading and trailing indicators, an AI model’s training accuracy is improved to 97.1\%. Comparatively, the accuracies reached 96.3\% and 89.3\% when the AI model was trained on only the leading and only the trailing indicators, respectively.

Overall, CAWFI includes all daily California wildfire activity from 2012 to 2018, daily meteorological and environmental data from 2012 to 2022, and the most recent geological data from August 2023. Wildfire risk and indicator data are represented by daily heatmaps, with brighter pixels representing higher readings at the pixel’s corresponding geographical location.

\section{Methods: Data Collection and Processing}
\label{sec:methods}
In addition to identifying the suitable indicators for the CAWFI dataset, it is also important to consider other requirements for data collection. Firstly, the historical wildfire data must span over a continuous period of at least a few years and specify the location and date of each wildfire activity. Secondly, all indicator data must come from publicly accessible remote sensor networks covering the same time period and the corresponding geographical area as the historical wildfire occurrence data. Additionally, the time interval between consecutive data points should be no longer than a day with geological indicator data as an exception. 

\begin{figure}[t]
  \centering
  \includegraphics[width=1.0\linewidth]{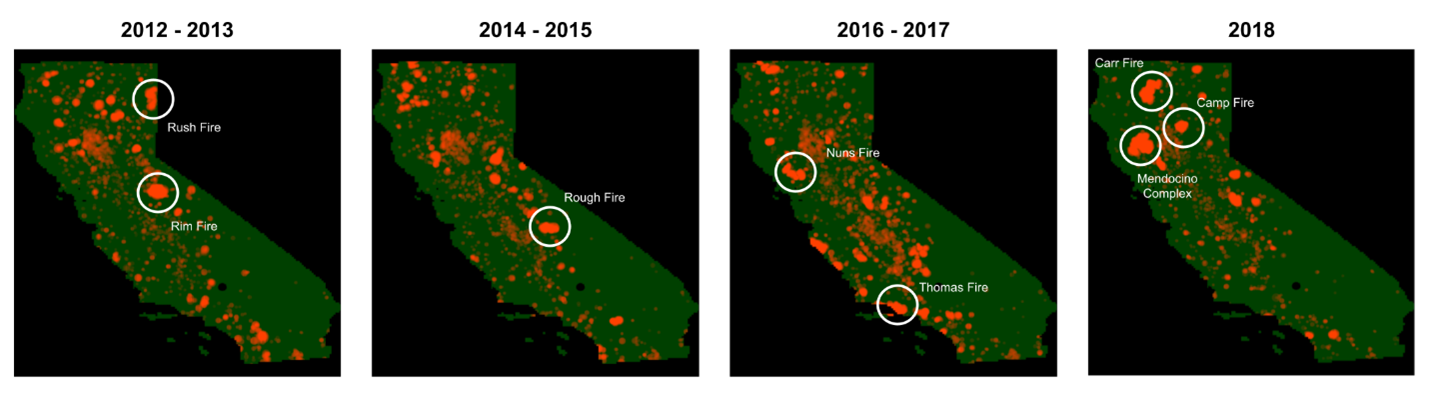}
  \caption{Spatial distributions of wildfires across California, with significant wildfires such as the Rush Fire and Rim Fire of 2012-2013, Rough Fire of 2014-2015, Nuns Fire and Thomas Fire for 2016-2017, and Carr Fire, Camp Fire, and Mendocino Complex for 2018. The visualized data circled in white show concentrated areas of high wildfire activity in red, reinforcing the reliability of the FRP data in reflecting real-life wildfire activity and historical disasters.}
  \label{fig:2}
\end{figure}

The wildfire incidence data in CAWFI originates from WildfireDB, an open-source dataset of 17.8 million data points cataloging wildfire activity within California from 2012 to 2018, inclusive \citep{Singla2020}. WildfireDB contains daily information including location (specifically, which of the 375 × 375 m$^2$ polygons that WildfireDB creates from partitioning California), date, and Fire Radiative Power (FRP) measurements. FRP measures the rate of infrared radiation emitted and indicates the rate of vegetation being burned and the severity of the wildfire at abnormally high values. The 375 × 375 m$^2$ polygons are sorted by day and arranged by location to construct daily images of wildfire heatmaps. Specifically, wildfires are depicted on a heatmap (originally all zeros) as a circle of ones, with zeros representing no wildfire and ones representing wildfires. The radius of a circle is directly proportional to the fourth root of the FRP reading according to the Stefan-Boltzmann law (which states that P $\propto$ T$^4$) to prevent wildfires with smaller FRP values from being negligible. Figure ~\ref{fig:2} depicts the resulting heatmaps aggregated over different years and shows how they reflect California’s most devastating wildfires within the corresponding time periods. 

\section{Discussion}
\label{sec:discussion}
The presented CAWFI dataset should be widely applicable for wildfire prediction within California. It is also envisioned to set a standard for wildfire and government agencies planning on implementing wildfire prediction solutions elsewhere. For these purposes, the raw wildfire and indicator data and the Python code necessary to process and reconstruct CAWFI are provided at https://github.com/rohan-tan-bhowmik/CAWFI-Data. The database provides all mentioned meteorological, environmental, and wildfire data in both CSV and SQL files as well the necessary Python files and corresponding instructions needed to reconstruct and visualize heatmap data. Additionally, images and/or retrieval instructions are provided for geological and ecological indicators.

\begin{figure}[t]
  \centering
  \includegraphics[width=1.0\linewidth]{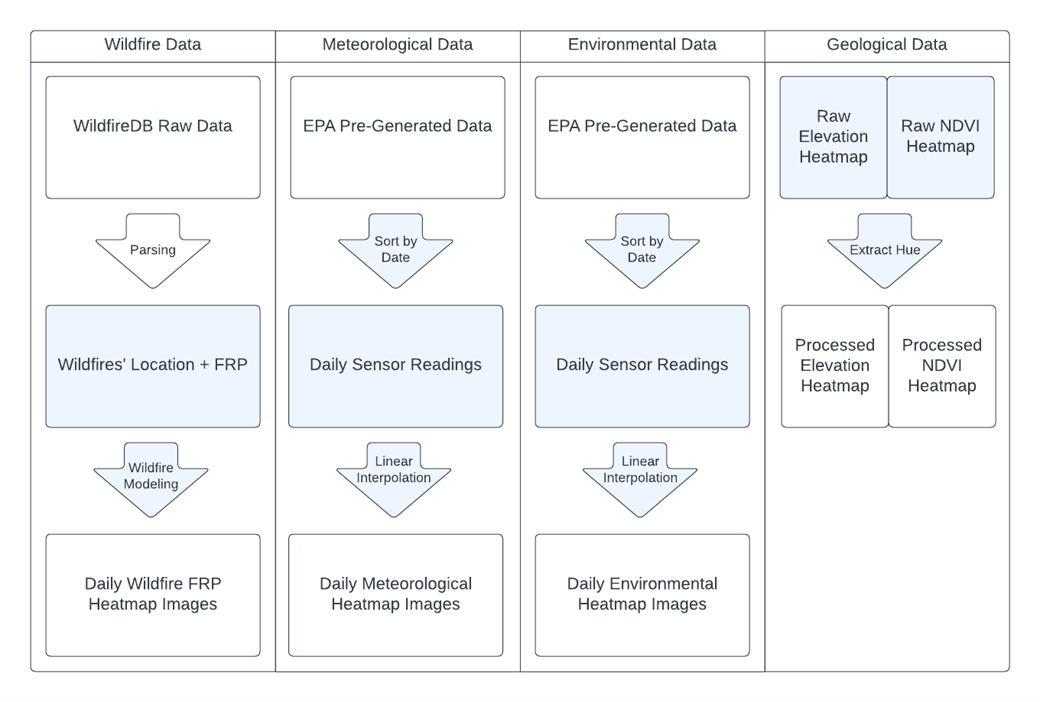}
  \caption{Segmented flow chart detailing the data conversion process steps used in the wildfire prediction paper from Bhowmik et al \citep{Bhowmik2023} Raw wildfire data is first parsed for date, location, and FRP data and then added as spots on heatmaps with varying radius based on wildfire severity. Raw meteorological and environmental data are sorted by date and then processed through linear interpolation to create heatmaps. Raw elevation and NDVI geological data has hue data extracted to form heatmaps, while ecological data is available via LandFire. All components in blue have been made available alongside CAWFI at https://github.com/rohan-tan-bhowmik/CAWFI-Data.}
  \label{fig:3}
\end{figure}

The efficacy CAWFI has already been demonstrated by a wildfire prediction technique utilizing the dataset \citep{Bhowmik2023}. In that study, a novel U-Convolutional-LSTM (Long Short-Term Memory) neural network was developed specifically for the task of predicting wildfires. The model architecture was designed to extract key spatial and temporal features within the meteorological, environmental, and geological data. Figure ~\ref{fig:3} details the pipeline from raw wildfire and indicator data to processed heatmaps, including the involved data files and processing scripts included in CAWFI corresponding to each step. As described in the previous study, wildfire risk in an area was defined as the model’s confidence (as a percentage) of the corresponding pixel on the output risk map; accuracy was defined as the model outputs’ proximity to the true wildfire risk heatmaps.

When trained on historical wildfire, meteorological, and environmental data from 2012 to 2017, the network achieved up to 97\% testing accuracy for up to 2 weeks in advance of the largest wildfires of the 2018 wildfire season in California. In comparison, traditional Convolutional Neural Network (CNN) techniques scored 76\%. Here, the method used to calculate accuracy measures the distance in value between the actual and predicted wildfire heatmap. When geological data was implemented as well and the U-Convolutional-LSTM network was retrained on all data from 2012 to 2017, the technique successfully predicted 85.7\% of wildfires larger than 300,000 acres in area. This successful implementation of wildfire prediction demonstrates the efficacy of CAWFI for use over California. 

CAWFI’s coverage is currently limited to California. While historical and real-time meteorological, environmental, and geological data are accessible through government sites and other databases nationwide, comprehensive and well-annotated historical wildfire data is not available to the public. A potential solution to this issue could be to train a wildfire prediction model without the context of location. If an AI network can be trained to predict wildfires within random areas of California, the same approach could then potentially be applied to any other area in the world. A model developed and trained based on California data would have limited success in other regions with different climate patterns such as Alaska and Canada. As climate change and disruptions of ecosystems continue to exacerbate severe wildfires, it is of paramount importance for countries to implement sensor networks and make their wildfire data openly accessible to the public.

\section{Conclusions and Future Outlook}
\label{sec:conclusions}
This work presents a wildfire prediction database, CAWFI, developed for training and testing wildfire prediction techniques. Specifically, CAWFI expanded upon existing wildfire datasets with historical wildfire data in California to include additional meteorological, environmental, and geological data as indicators that impact wildfire inception and spread. The efficacy of CAWFI has been demonstrated through the successful implementation of a California wildfire prediction solution utilizing a novel spatio-temporal neural network architecture. All raw data and code for reconstructing CAWFI is provided for research groups to develop such solutions and create similar databases for other regions.

In the future, extending CAWFI to accommodate a longer historical record of wildfire data and an overall larger area of coverage would expand its impact. It may also serve as a framework for wildfire control and government agencies to build the necessary infrastructure and gather these pertinent indicators.

With this work, we aim to define a standard for wildfire prediction solutions in the future. This will provide fire authorities with the ability to distinguish which wildfires would pose serious environmental and social threats in advance. Combined with firefighting technology and public alert systems, AI models trained on such data would be able to help quell potential megafires and spread awareness during high wildfire risk, thereby minimizing unjustified money, large infrastructure, and any lives lost due to wildfires globally.

\section*{Author Contributions}
Conceptualization, R.T.B.; methodology, R.T.B., Y.S.J., and J.A.; software, R.T.B.; validation, R.T.B., Y.S.J., and J.A.; formal analysis, R.T.B., Y.S.J., and J.A.; investigation, R.T.B., Y.S.J., J.A., M.P., and K.N.; resources, R.T.B., M.P., and K.N.; data curation, R.T.B.; writing---original draft preparation, R.T.B.; writing---review and editing, Y.S.J., J.A., M.P., and K.N.; visualization, R.T.B.; supervision, K.N.; project administration, R.T.B.; funding acquisition, N/A. All authors have read and agreed to the published version.

\section*{Data Availability Statement}
All relevant data created from this study is available at: \url{https://github.com/rohan-tan-bhowmik/CAWFI-Data}.

\section*{Acknowledgments}
We’d like to thank Singla et al. for helping provide and assist with wildfire data collection via WildfireDB, the Environmental Protection Agency for environmental and meteorological indicators, and various government agencies for geological indicators.



\begingroup
\sloppy
\bibliography{references}
\endgroup

\end{document}